# Single Shot Active Learning using Pseudo Annotators

Yazhou Yang                                                                              Y.YANG-4@TUDELFT.NL
Marco Loog                                                                                M.LOOG@TUDELFT.NL
Pattern Recognition Laboratory, Delft University of Technology, The Netherlands

## Abstract

Standard myopic active learning assumes that human annotations are always obtainable whenever new samples are selected. This, however, is unrealistic in many real-world applications where human experts are not readily available at all times. In this paper, we consider the single shot setting: all the required samples should be chosen in a single shot and no human annotation can be exploited during the selection process. We propose a new method, Active Learning through Random Labeling (ALRL), which substitutes single human annotator for multiple, what we will refer to as, pseudo annotators. These pseudo annotators always provide uniform and random labels whenever new unlabeled samples are queried. This random labeling enables standard active learning algorithms to also exhibit the exploratory behavior needed for single shot active learning. The exploratory behavior is further enhanced by selecting the most representative sample via minimizing nearest neighbor distance between unlabeled samples and queried samples. Experiments on real-world datasets demonstrate that the proposed method outperforms several state-of-the-art approaches.

## 1. Introduction

In many machine learning applications, the availability of a large amount of data offers opportunities to boost the prediction performance. Even if data is abundant, a major issue remaining is that labeling the data is usually time-consuming and expensive. For example, it is costly to hire many dermatologists to annotate the 129,450 clinical images of skin cancer used in (Esteva et al., 2017). Active learning, which iteratively selects the most informative samples and queries the labels from human experts, has demonstrated its ability to reduce the annotation cost and maintain good learning performance in various applications (Settles, 2010; Wang & Hua, 2011).

The strength of active learning in reducing annotation cost stems from the fact that it can iteratively query its preferred unlabelled examples for labelling and simultaneously update its selection strategy according to the feedback from a human expert. Indeed, conventional active learning assumes a human in the loop such that it can iteratively learn from the received label information. This also implies that in the classical setting of active learning, human annotators should be always readily available for labeling whenever new unlabeled samples are queried. However, this assumption may not hold in some real-world applications since (1) human annotator is unlikely to be present at all time, e.g. human annotator may get tired or need a rest, (2) and active learning process has to be suspended until the annotator reappear.

To mitigate the issue of human annotators being missing in the loop, we consider a single shot setting of pool-based active learning, where few labeled samples and a potentially large number of unlabeled instances are available and the active learner is asked to choose a query set $\mathcal{Q}$ in a single shot (Contardo et al., 2017). Simply using standard myopic active learning algorithms to select the top-ranked samples is not a good choice since it fails to consider the redundancy among these top instances (Settles, 2010). Figure 1a shows an example of the failure of standard uncertainty sampling (Lewis & Gale, 1994; Settles, 2010). We can observe that in the single shot scenario, uncertainty sampling chooses a subset of samples which overlap each other. And it fails to explore the other two clusters.

In this paper, we concentrate on adapting standard active learning algorithms, which need real label information for exploitation, to the single shot setting. We propose a new method, called Active Learning through Random Labeling (ALRL). Our method introduces multiple annotators, which we refer to as pseudo annotators, to take the place of the human expert used in active learning cycle. The pseudo annotators are independent from each other and present uniformly random labels whenever new unlabeled samples are queried. Even though the pseudo annotators do not add any information telling us anything about the true labels,

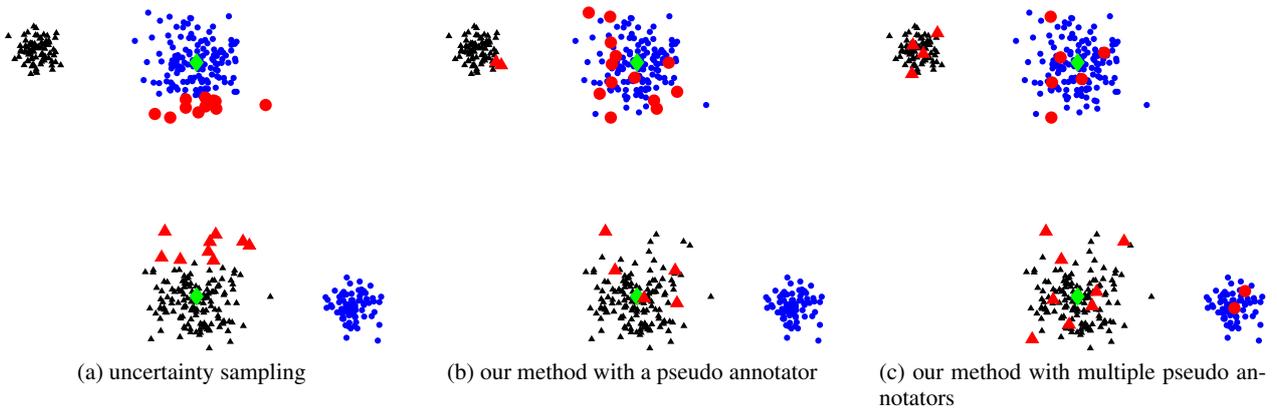

(a) uncertainty sampling    (b) our method with a pseudo annotator    (c) our method with multiple pseudo annotators

*Figure 1.* An illustration of different active learning algorithms in the single shot setting. These figures show the queried points chosen by (a) standard uncertainty sampling (i.e. maximum entropy (Lewis & Gale, 1994; Settles, 2010)), (b) uncertainty sampling + random labeling (with a single pseudo annotator) and (c) uncertainty sampling + random labeling (with multiple pseudo annotators). Blue and black points represent two different classes. Two green diamond points are initial labeled instances while the red points are queried samples.

regular active learning methods can still benefit from receiving such random labels. The improvement comes from the exploration ability provided by this random labeling mechanism. As we will see, the exploratory behavior is further enhanced by fine-tuning the results that come from multiple pseudo annotators by selecting the sample that well-represents the unlabeled data. The proposed method is a general approach, which can be incorporated with both simple active learners, e.g. uncertainty sampling (Lewis & Gale, 1994) and sophisticated ones, e.g. variance maximization (Yang & Loog, 2018). We show the efficiency of our method on real-world datasets, in comparison with state-of-the-art approaches.

## 2. Related Work

A brief review of the work related to our single shot active learning is given, including myopic and batch mode active learning, optimal experimental design, data subset selection, and single shot selection.

**Myopic vs. batch**. Active learning can be roughly divided into two categories according to the number of queried samples at a time (Yang & Loog, 2016). The first one is myopic active learning, where only a single instance is selected in each iteration. Many well-known algorithms, such as query-by-committee (QBC) (Seung et al., 1992), uncertainty sampling (Lewis & Gale, 1994; Tong & Koller, 2002), error reduction (Roy & Mccallum, 2001), maximum model change (Freytag et al., 2014), variance reduction (Schein & Ungar, 2007), and variance maximization (Yang & Loog, 2018) belong to this group.

The second one is batch mode active learning, where a batch of samples is selected simultaneously (Hoi et al., 2006; Guo & Schuurmans, 2007; Chakraborty et al., 2013; 2015). Conventional batch mode active learning methods first select a fixed number of unlabeled instances and then ask for the real labels from human experts. Subsequently, they make use of the received real label information to update their selection criteria and continue choosing the next round of a group of unlabeled samples. When the batch size is very large, e.g. the number of required samples in total, batch mode setting is transformed into the considered single shot case. In other words, single shot active learning can be viewed as a particular case of batch mode active learning with the batch size being equal to the sampling budget. However, Brinker (2003) found that it is preferable to set batch size as small as possible, with the possible reason that the selection criterion can be updated more frequently if batch size is small. This implies that directly using batch mode active learner for one single shot setting may lead to a decrease of the learning performance.

**Optimal experimental design**. There also exist some active learning methods which do not require true labels for samples selection at all, such as transductive experimental design (TED) (Yu et al., 2006) and graph-based variance minimization methods (Ji & Han, 2012; Ma et al., 2013). These approaches usually attempt to minimize the expected variance of a statistical model, where the label information is omitted in the calculation of such variance. Hence, they are well matched with this single shot setting since no human annotation is needed during the selection process. However, since they do not utilize label information, they mainly focus on selecting representative samples and do not make use of label information even when some labels of samples are available.

**Data subset selection**. Data subset selection is also closely related to our single shot setting since both of them aim to select an informative subset. The subset selection problem has been well studied in the literature. When the input samples are feature vectors, many works focus on finding representative samples by searching in low-dimensional subspace (Elhamifar et al., 2012; Boutsidis et al., 2009) or using some clustering algorithms, e.g. $k$-means. Other efforts have been devoted to finding the representatives by using pairwise similarities between data points (Frey & Dueck, 2007; Elhamifar et al., 2016). For example, Elhamifar et al. (2016) proposed a dissimilarity-based sparse subset selection (DS3) method to minimize the difference between source data and target data. However, some methods, e.g. DS3 and the work in (Frey & Dueck, 2007), cannot exactly determine the number of selected representative points beforehand.

**Single shot selection**. Few efforts have been devoted to single shot active learning problem. Contardo et al. (2017) combined meta-learning and active learning to learn an active learning strategy. After the selection strategy is learned, all the required samples are queried in a single shot. However, their method required additional supervised data to train their model, which is not realistic in many active learning applications since only little or even no labeled data is available. Our proposed method does not require extra supervision information and also can work without any labeled data.

## 3. Active Learning using Random Labeling

This section presents our novel approach for querying an informative subset for human annotation in a single shot in detail. The proposed method offers an alternative view of the subset selection problem, which adapts standard myopic active learners to the single shot setting through random labeling.

### 3.1. Motivation

Many active learning algorithms, which make use of label information for samples selection, focus on exploitation by querying samples near the decision boundary to refine the classification model, e.g. uncertainty sampling. On the contrary, some other methods concentrate on the exploration by selecting the most representatives of the unlabeled instances. Most of these approaches, e.g. Transductive experimental design (Yu et al., 2006) and Hessian optimal design (Lu et al., 2011), do not use the class information of selected samples at all.

In the single shot setting, in the absence of experts annotations, regular exploitation-based active learning algorithms may fail because they cannot update their exploitation criterion for every sample selected. Conducting the exploitation on the basis of initial training data without further updating the selection criterion is likely to mislead the active learner to select uninformative and redundant samples. As is illustrated in Figure 1a, uncertainty sampling, which concentrates on exploitation, queries less informative samples in a single shot. And there are high redundancies among these queried samples. It shows that, in general, pure exploitation without subsequent updating can indeed be harmful in the single shot setting.

To overcome the disadvantage of pure exploitation, we enable standard myopic active learners to explore by using multiple of our so-called pseudo annotators. The value of employing such pseudo-annotators will be explained in Subsection 3.2. First, however, we explain more precisely what a pseudo-annotator does. A pseudo annotator does not know anything about the true labels and just randomly guesses a class category when annotating an unlabeled instance. For example, given that $C = \{c_1, c_2, \ldots, c_p\}$ is the set of possible labels, the pseudo annotator randomly and uniformly selects one label $c_i$ from $C$ for each queried unlabeled instance. This also implies that the randomly assigned labels of different unlabeled samples are totally independent from each other.

### 3.2. The Proposed Method: Random Labeling

We start with the basic setting of single shot active learning. We have relatively little labeled data $\mathcal{L} = \{(x_i, y_i)\}_{i=1}^{n_l}$, where $x_i \in \mathbb{R}^d$ is a feature vector and $y_i$ is the label of $x_i$. In addition, a large pool of unlabeled examples $\mathcal{U}$ is also available. The task is to select a budget of $N$ samples from the unlabeled pool $\mathcal{U}$ in a single shot. When $N$ instances are determined, they are categorized by human annotators and added to the labeled data $\mathcal{L}$.

Figure 2 shows the flow chart of our proposed method. $\mathcal{Q}$ denotes the already selected but still unlabeled data. Our method chooses $N$ samples in a sequential way, which means only one sample is queried at a time.

To start with, we set $\mathcal{Q} = \emptyset$ and select the first instance according to the myopic active learner $\mathcal{A}$ trained on $\mathcal{L}$. From then on, the instances in $\mathcal{Q}$ are randomly labeled by the pseudo annotators and used to select the next sample. Trained on correctly labeled $\mathcal{L}$ and randomly labeled $\mathcal{Q}$, the active learner we use is able to explore larger regions than that only trained on $\mathcal{L}$. As shown in Figure 1b, uncertainty sampling is still used as the active learner but our method can make it select diverse samples instead of purely those samples lying close to initial decision boundary. This verifies that random labeling can indeed help exploration. Note that in each iteration, all the samples in $\mathcal{Q}$ are relabeled by the pseudo annotators, which means that the assigned la-bels may be different from that obtained in the last round.

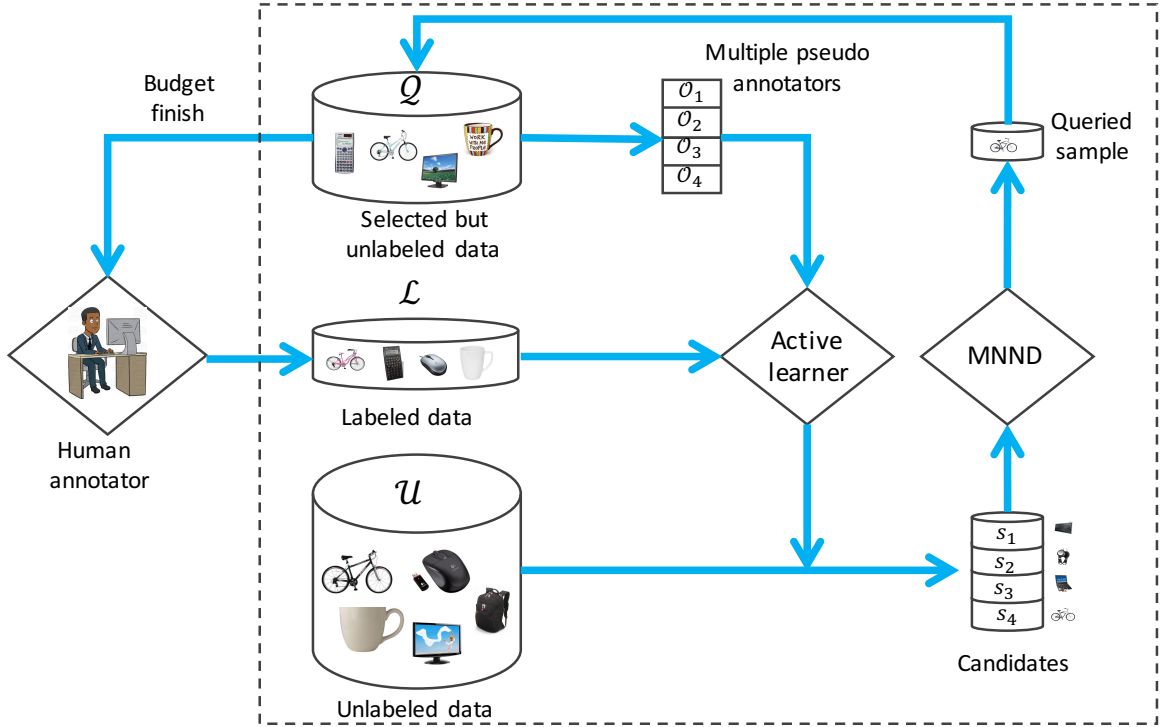

*Figure 2.* Schematic illustration of the proposed method. $\mathcal{L}$ and $\mathcal{U}$ denote the truly labeled data and unsupervised data, respectively. $\mathcal{Q}$ represents those already selected but still unlabeled data. $\{\mathcal{O}_1, \mathcal{O}_2, \ldots, \mathcal{O}_m\}$ ($m = 4$ in this figure) denotes a set of pseudo annotators where each pseudo annotator $\mathcal{O}_i$ always randomly and uniformly label these samples in $\mathcal{Q}$. Only when budget is finished will the samples in $\mathcal{Q}$ be labeled by human annotator. $\{s_1, \ldots, s_m\}$ are the candidates selected by active learner based on each pseudo annotator. MNND stands for minimizing nearest neighbor distance between unlabeled data and already queried data. In the end, the most representative sample from $\{s_1, \ldots, s_m\}$ is chosen by MNND and is added to $\mathcal{Q}$.

A major difference between our random labeling and standard active learning algorithms is that method employs multiple pseudo annotators whereas classical active learning assumes that only a single annotator is available. The motivation of using multiple pseudo annotators is that a single random labeling strategy would make the output of our algorithm too depend on the quality of randomly assigned labels. It could happen that our method unfortunately queries an uninformative sample because of the poor random labels. To tackle this problem, we simply decide to use $m$ different pseudo annotators, i.e. $\{\mathcal{O}_1, \mathcal{O}_2, \ldots, \mathcal{O}_m\}$, and fine-tune the result by selecting the most representative instance obtained by using multiple pseudo annotators. Note that these pseudo annotators are independent from each other.

More specifically, we first use $\mathcal{O}_i$ to randomly label all the samples in $\mathcal{Q}$ and then apply an active learner $\mathcal{A}$ to choose one candidate $s_i$ from unlabeled data $\mathcal{U}$ based on truly labeled $\mathcal{L}$ and randomly labeled $\mathcal{Q}$. We repeat this procedure $m$ times with different pseudo annotators, result in obtaining $m$ different candidates, i.e. $\{s_1, s_2, \ldots, s_m\}$, since each pseudo annotator is highly likely to present different random labels to $\mathcal{Q}$. Subsequently, we can evaluate the candidate samples and select the one which best represents the unlabeled samples.

Overall, as shown in Figure 2, the proposed random labeling mechanism uses a two-step strategy: it first employs multiple pseudo annotators to impel regular active learner to explore and choose an informative candidate set; then the most representative sample is queried from the candidate set. The overall training procedure of our method is given in Algorithm 1.

In this work, we consider minimizing the overall nearest neighbor distance between unlabeled data and queried data to choose the most representative sample. We call this technique Minimizing Nearest Neighbor Distance (MNND for short). We will further explain why and how to implement the MNND in Subsection 3.3.

**Algorithm 1** Active Learning with Random Labeling
**Require:** Labeled data $\mathcal{L}$, unlabeled data $\mathcal{U}$, subset $\mathcal{Q} = \emptyset$, Active Learner $\mathcal{A}$, pseudo annotators $\{\mathcal{O}_1, \mathcal{O}_2, \ldots, \mathcal{O}_m\}$
1: **repeat**
2:    **for** $i = 1$ **to** $m$ **do**
3:       Samples in subset $\mathcal{Q}$ are randomly and uniformly labeled by the pseudo annotator $\mathcal{O}_i$;
4:       Train on $\mathcal{L} \cup \mathcal{Q}$ and use active learner $\mathcal{A}$ to select the most informative sample denoted by $s_i$;
5:    **end for**
6:    Select the sample $x^*$ from $\{s_1, s_2, \ldots, s_m\}$ by using MNND (see Equation 2);
7:    update $\mathcal{Q} \leftarrow \mathcal{Q} \cup \{x^*\}, \mathcal{U} \leftarrow \mathcal{U} \backslash \{x^*\}$;
8: **until** Budget is reached
9: Human expert annotators the instances in $\mathcal{Q}$ with true labels $Y_\mathcal{Q}$, update $\mathcal{L} \leftarrow \mathcal{L} \cup \{\mathcal{Q}, Y_\mathcal{Q}\}$;

### 3.3. Minimizing Nearest Neighbor Distance

Figure 3 illustrates the main idea behind minimizing nearest neighbor distance between unlabeled samples and queried samples. The yellow dots represent the unlabeled samples, e.g. samples in $\mathcal{U}$ whereas the red squares stand for these already chosen samples, e.g. samples in $\mathcal{L} \cup \mathcal{Q}$. The red dot indicates that this instance is chosen as the next queried data point, followed by a calculation of the minimum nearest neighbor distance. First, we define the overall nearest neighbor distance between unlabeled data $\mathcal{U}$ and the remaining already chosen data which includes both labeled data $\mathcal{L}$ and $\mathcal{Q}$ as follows:

$$Dis(\mathcal{U}, \mathcal{L}, \mathcal{Q}) = \sum_{u \in \mathcal{U}} \min_{x \in \mathcal{L} \cup \mathcal{Q}} \|u - x\| \qquad (1)$$

where $\|u - x\|$ denotes the Euclidean distance between unlabeled data point $u$ and labeled (in our case possibly randomly labeled) data point $x$.

For example, in Figure 3, for all unlabeled samples, we first find their nearest neighbor from $\mathcal{L} \cup \mathcal{Q}$ and then sum over all the pair distances between unlabeled data points and their corresponding nearest neighbors. Finding such nearest neighbor can be interpreted in two ways: (1) it can be seen as classifying unlabeled samples using 1-nearest neighbor algorithm; (2) it can also viewed as clustering unlabeled instances according to these already chosen data points, where each instance in $\mathcal{L} \cup \mathcal{Q}$ is considered as the cluster centroid. If $\mathcal{U}$ is not empty and $Dis(\mathcal{U}, \mathcal{L}, \mathcal{Q})$ reaches its minimum value 0, it implies that all unlabeled data points can find some samples which are exactly the same as themselves. This also indicates that all unlabeled data can be perfectly classified by the 1-nearest neighbor algorithm. Therefore, $Dis(\mathcal{U}, \mathcal{L}, \mathcal{Q})$ can be considered as a measure of how well the labeled data can represent the unlabeled data. The smaller the value of $Dis(\mathcal{U}, \mathcal{L}, \mathcal{Q})$, the more representative of already queried samples.

Now let us return to how to select the most representative sample from these candidates obtained by employing multiple pseudo annotators. For example, assume that our method uses $m$ pseudo annotators and obtains $m$ candidates: $\mathcal{S} = \{s_1, \ldots, s_m\}$ with $s_i \in \mathcal{S}$. We prefer the sample $s$ which can lead to a minimum nearest neighbor distance once chosen as the next queried sample. The intuition behind is that the smaller the value of $Dis(\mathcal{U}, \mathcal{L}, \mathcal{Q} \cup s)$, the more representative the queried samples $s$ is. Therefore, we consider selecting $x^*$ to minimize the nearest neighbor distance between queried data and unlabeled data as follows:

$$x^* = \arg\min_{s \in \mathcal{S}} Dis(\mathcal{U}, \mathcal{L}, \mathcal{Q} \cup s) \qquad (2)$$

As we see, there are four candidates $\{s_1, s_2, s_3, s_4\}$ chosen by using four pseudo annotators in Figure 3. If $s_4$ is chosen as the next queried instance (shown in Figure 3(b)), it will result in a minimum nearest neighbor distance than that other samples are queried (e.g. as shown in Figure 3(a), selecting $s_1$ will lead to a larger value of $Dis(\mathcal{U}, \mathcal{L}, \mathcal{Q} \cup s)$). Finally, our algorithm chooses $s_4$ as the next queried instance. This also implies that our method prefers these samples which are located in the high density region.

To demonstrate the effectiveness of MNND, let us compare Figure 1b and Figure 1c. We observe that our method without using MNND fails to select the points in the bottom right corner (in Figure 1b). However, when MNND is utilized, our method can select samples from all four clusters and most of the queried points are in high density regions. As it turns out, MNND is able to further enhance the exploratory behavior of standard active learning algorithms.

## 4. Comparisons and Connections

This section discusses the connections and differences between our method and other relevant approaches.

To start with, we first describe the active learning techniques which are combined with our random labeling mechanism. Our method is generally designed for active learning algorithms which make use of label information for selection. We employ our method in combination with two active learning methods. The first one is an uncertainty sampling strategy, the maximum entropy criterion (MaxE for shot) (Settles, 2010). Though MaxE is quite simple, it performs well in comparison with many myopic active learning algorithms (Yang & Loog, 2016). The other one we consider is a recently proposed myopic active learning method called Maximizing Variance for Active Learning (MVAL for short). MVAL shares some similarities with the classical query-by-committee (Seung et al., 1992). MVAL forms a committee that consists of models trained on currently labeled data and each unlabeled sample with all possible labels. It trains each committee member and records the posterior probabilities of unlabeled samples to form so-called retraining information matrices (RIMs). These RIMs are used to compute the disagreement among each committee member on all unlabeled samples. More specifically, MVAL estimates two kinds of variance to evaluate the informativeness and representativeness and fuses these variances as a measure of the disagreement. Finally, MVAL queries the sample that causes maximum disagreement among all committee members. The main differences between MaxE and MVAL are that MaxE only concentrates on exploitation while MVAL considers both exploitation and exploration.

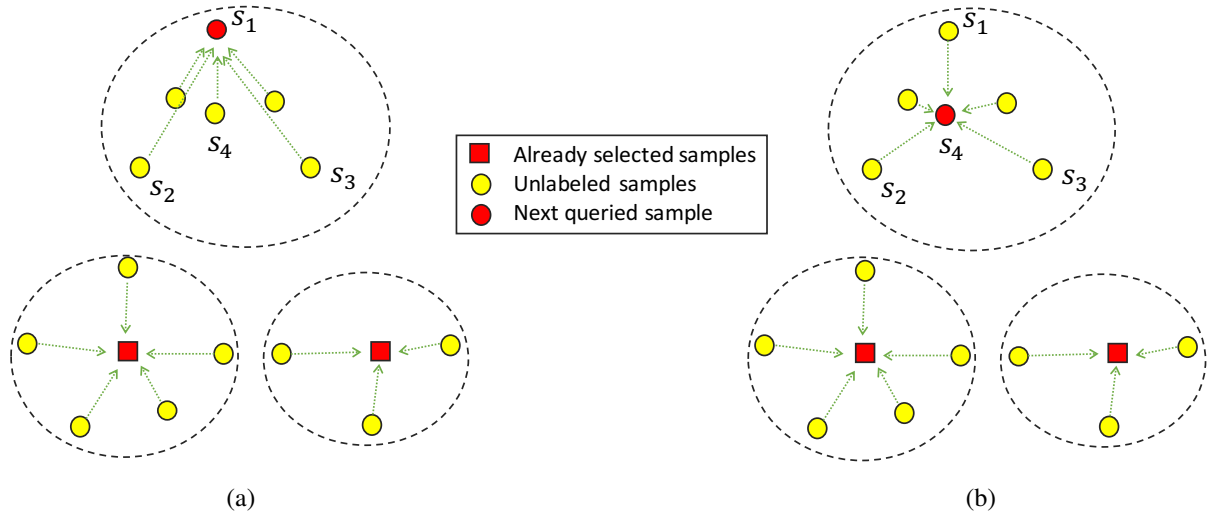

Figure 3. Illustration of minimizing nearest neighbor distance between unlabeled samples and queried samples. Arrows indicate the distance between unlabeled instances to their nearest already selected instances. The dashed ellipses indicates that these unlabeled samples inside share a common nearest neighbor. $\{s_1, s_2, s_3, s_4\}$ are the candidates chosen by using multiple pseudo annotators. (a) $s_1$ is assumed to be selected as the next queried sample; (b) $s_4$ is assumed to be selected as the next queried sample. Since choosing $s_4$ will lead to a smaller overall distance between unlabeled samples and queried samples, our algorithm chooses $s_4$ as the next queried instance.

Many efforts have been devoted to density or diversity based active learning (Brinker, 2003; Liu et al., 2008; Settles & Craven, 2008; Yang et al., 2015; Zhu et al., 2010). The common idea behind these is that we should select samples which are representative of unlabeled data, e.g. both Settles & Craven (2008) and Zhu et al. (2010) used the similarity between unlabeled samples as a measure of density and combined the density measure with the uncertainty measure. However, these methods may get into trouble in the single shot setting. The reason is that the uncertainty measure is fixed during the selection process since no true labels can be obtained and these methods may fail to balance density and uncertainty. Similar to these density-based approaches, our method also prefers querying representative samples by minimizing the nearest neighbor distance. For example, in Figure 3, our algorithm prefers selecting these samples which are close to the cluster centers. In this sense, the proposed method provides an alternative view to select representative samples.

The proposed method also has some connections with $k$-means++ algorithm (Arthur & Vassilvitskii, 2007). $k$-means++ uniformly and randomly chooses a data point as the first cluster center, and selects the next cluster center from remaining data points with probability proportional to their squared distance from the closest existing cluster center. After that, the chosen seeds are feed to start $k$-means clustering algorithm.

Four aspects distinguish our work from $k$-means++. To start with, the first point is chosen by the active learner $\mathcal{A}$ trained on initially labeled data $\mathcal{L}$. Secondly, the subsequent point is not chosen from all the remaining unlabeled samples. Our method only chooses the next queried point from a candidate set $S$ generated by multiple random annotators. Thirdly, our method determinately selects the sample which leads to a minimum neighbor distance once chosen while $k$-means++ randomly chooses the next sample with some kind of probability. Finally, our method does not use any subsequent clustering algorithms while $k$-means++ still needs use $k$-means clustering algorithm. In some sense, the proposed method can also be viewed as a sequentially adaptive clustering approach. The benefit of our method over $k$-means++ is that our method can make use of some existing supervised information, e.g. training active learner $\mathcal{A}$ on initially labeled data $\mathcal{L}$, while $k$-means++ is a pure unsupervised clustering approach.

The strength of the proposed method over random sampling is that our method considers both exploitation, i.e. training on the truly labeled data $\mathcal{L}$, and exploration, i.e. using randomly labeled $\mathcal{Q}$. More important is that we also use MNND to select representative instances. On the contrary, random sampling only does pure exploration and fails to consider the representativeness of unlabeled samples. One work (Li et al., 2015) also proposed to use some randomly selected samples to explore. However, that technique is particularly designed for binary classification tasks and it is unclear how to extend it to multi-class classification.

## 5. Experiments

We test the empirical performance of the proposed method and compare it against other subset selection approaches. Additional comparisons are also made between our method and conventional batch mode active learning algorithms. We first describe the used test datasets and experimental setup, followed by an analysis of the experimental results.

### 5.1. Datasets

We use 10 benchmark datasets in our experiments, some of which are image classification tasks, such as two handwritten digit datasets, the MNIST (LeCun et al., 1998) and the USPS dataset (Hull, 1994). For the MNIST and the USPS dataset, the gray-scale pixel values are used as the features. We also use three pre-processed subsets in the *Office* dataset (Gong et al., 2012), including the Amazon, Webcam and Caltech datasets. These three sets contain 10 common classes which are from different sources, e.g. the Amazon dataset contains images downloaded from online merchants and the Webcam set uses low-resolution images by a web camera. SURF features are firstly extracted and then encoded into an 800-bin histograms. In addition, we also experiment on five standard datasets (Fang et al., 2013), including five categories of images taken from Caltech101 (C), ImageNet (I), LabelMe (L), SUN09 (S), VOC2007 (V). The selected five categories are: bird, car, chair, dog and person. Following (Fang et al., 2013), we use the pre-extracted DeCAF6 features. For computational efficiency, sub-sampling and principal component analysis (PCA) are applied on some of the larger datasets to reduce the sample size and feature dimensionality. The detailed information of these preprocessed test datasets after preprocessing is listed in Table 1.

Table 1. Characteristics of the preprocessed test datasets: the number of instances (#n), the feature dimensionality (#fea) and the number of class (#c). Refer to the text in the beginning of Subsection 5.1 to see what C, L, V, I, and S stand for.

| Dataset | (#n, #fea, #c) | Dataset | (#n, #fea, #c) |
|---|---|---|---|
| C | (467, 4096, 5) | I | (500, 4096, 5) |
| L | (410, 4096, 5) | S | (350, 4096, 5) |
| V | (500, 4096, 5) | Amazon | (500, 50, 10) |
| Webcam | (295, 50, 10) | Caltech | (500, 50, 10) |
| MNIST | (1000, 60, 10) | USPS | (1000, 60, 10) |

### 5.2. Experimental Setup

In our experiments, each dataset is randomly and repeatedly divided into training and test sets of equal size. We randomly select one sample from each class as the initial labeled set. All the experiments are repeated 20 times and the average performances are reported. The number of queried samples varies from $\{20, 40, 60, 80, 100, 120\}$ in our experiment.

We compare the proposed method, Active Learning through Random Labeling (ALRL for short), with random sampling and the following algorithms:

- USDM: Uncertainty sampling with diversity maximization, which retains the uncertainty and maximizes the diversity simultaneously (Yang et al., 2015).

- BatchRank: It balances the informativeness and diversity and offers some relaxations to solve the optimization problem (Chakraborty et al., 2015).

- $k$-means++: It applies $k$-means++ (Arthur & Vassilvitskii, 2007) algorithm and selects the sample nearest to the centroid of a cluster.

- TED: Transductive experimental design chooses examples to minimize the variance of ridge regression model (Yu et al., 2006).

- $V$-opt: It selects samples to minimize the $V$-optimality on Gaussian Random Fields (GRFs) (Ji & Han, 2012).

- $\Sigma$-opt: Similar to $V$-opt, it minimizes the $\Sigma$-optimality on GRFs (Ma et al., 2013).

- DS3: It selects representative samples by minimizing the dissimilarity between selected data and the remaining data (Elhamifar et al., 2016).

Among these compared methods, two methods, USDM and BatchRank, are the most recent state-of-the-art batch mode active learning algorithms. Since $k$-means++ is relatively sensitive to the initialization seeds, we repeat it 500 times with different random seeds and choose the one which performs the best based on the objective function of $k$-means++. For fairness, linear SVM with the same parameter setting is used to evaluate the performances of all compared algorithms. We use the LIBSVM package (Chang & Lin, 2011) and empirically set the regularization parameter $C = 10$. The number of pseudo annotators $m$ is empirically set 10 in all experiments.

### 5.3. Results

We first investigate whether the proposed random labeling mechanism can help improve the performance of standard active learning algorithms. Subsequently, we compare our method with other subset selection approaches.

#### 5.3.1. THE EFFICIENCY OF RANDOM LABELING

We show the average performance over all test sets of our method with MaxE and MVAL in Figure 4. ALRL_MaxE denotes the combination of MaxE and our method with a default $m$=10 pseudo annotators. We also show the performance of ALRL_MaxE ($m$=1) in which a single random annotator is used such that MNND does not play a role. It is the same setting with ALRL_MVAL and ALRL_MVAL ($m$=1). MaxE_True and MVAL_True refer to the two active learners that are obtained of the true is obtained fro a human ob server in every iteration. This is, in a sense, the best one can do and therefore serves as a natural upper bound for the performance.

It is obvious that standard myopic active learning algorithms perform poorly in the single shot setting, e.g. MaxE and MVAL are significantly worse than random sampling. The reason is that these active learners are likely to select samples which extensively overlap each other. As shown in Fig 1a, standard uncertainty sampling keeps selecting instances which are close to initial decision boundary and have considerable overlapping within each other. And it fails to query new data points from the other two clusters. From another perspective, this also implies that the key advantage of active learning over random sampling is that active learner can iteratively learn from the labels obtained from human annotator. In the single shot scenario where standard myopic active learners cannot query human annotator for labels and cannot update their selection criteria, it does make sense that these approaches demonstrate poor performance.

However, by adopting our proposed random labeling, both ALRL_MaxE ($m$=1) and ALRL_MVAL ($m$=1) outperform the original active learners (MaxE and MVAL) and random sampling. As shown in Fig 1b, our random labeling strategy impels uncertainty sampling to explore and select diverse data points without large overlapping. This demonstrates that in the single shot scenario, our random labeling mechanism can indeed boost the performance of regular active learners by promoting exploration.

We also investigate the benefit of our proposed MNND criterion. In Figure 4, ALRL_MaxE and ALRL_MVAL outperform their competitors, ALRL_MaxE ($m$=1) and ALRL_MVAL ($m$=1), respectively. This confirms the advantage of selecting representative samples by minimizing nearest neighbor distance. It also means that we can still expect better performance when multiple pseudo annotators are employed and the most representative sample are chosen by using MNND. We also observe that two active learners that received human feedback in every iteration, MaxE_True and MVAL_True, obtain the best performances. And our method comes very close to these two active learners. This demonstrates that even in the single shot setting, our proposed random labeling mechanism can produce promising results relatively comparable to that of active learner in the standard setting where human annotation is obtainable in each iteration.

#### 5.3.2. THE INFLUENCE OF MYOPIC ACTIVE LEARNER

Figure 5 illustrates the influence of myopic active learner chosen in our method. ALRL_Random means that random sampling is used as the active learner $\mathcal{A}$ in Algorithm 1. In addition, we also compare with Simple_MNND in which no pseudo annotators are used and the most representative sample is directly chosen from all the remaining unlabeled samples by using MNND. The difference between ALRL_Random and Simple_MNND is that ALRL_Random conducts MNND on a random subset while Simple_MNND implements MNND on all remaining unlabeled data. We can see that ALRL_Random and Simple_MNND perform similarly to each other. And both of them are surpassed by ALRL_MaxE and ALRL_MVAL. This implies that the exploitation produced by MaxE and MVAL increases the learning performance. Overall, ALRL_MVAL demonstrates the best performance. In the following experiments, ALRL refers to ALRL_MVAL, unless otherwise specified.

#### 5.3.3. PERFORMANCE COMPARISON WITH STATE-OF-THE-ART ALGORITHMS

Figure 6 compares our method ALRL to several state-of-the-art algorithms over 10 test datasets. We can see that our method obtains the best performance on most datasets, such as MNIST, USPS, Caltech, Amazon, L and Webcam. ALRL also demonstrate excellent performance on the I and V datasets, ranking in the second place. It only fails to remain among the top two methods on the C and S datasets. TED also shows good results on several datasets, especially on the V and I, on which it achieve the highest accuracy. Most of the remaining methods can perform quite well only on one or two datasets, e.g. DS3 exceed other methods on the C dataset. However, these two batch mode active learning algorithms USDM and BatchRank perform poorly on most datasets, e.g. BatchRank obtains worse performance than random sampling on 7 datasets. Incidentally, this supports our claim that standard batch mode active learning algorithms are likely to under-perform in the single shot setting. We show the average performance of the different algorithms in Figure 7. Our method achieves the best overall performance, with TED as a good second. $k$-means++ shows performance similar to TED, but under-performs when the budget is very small. Our method shows a clear advantage both with small and large budgets.

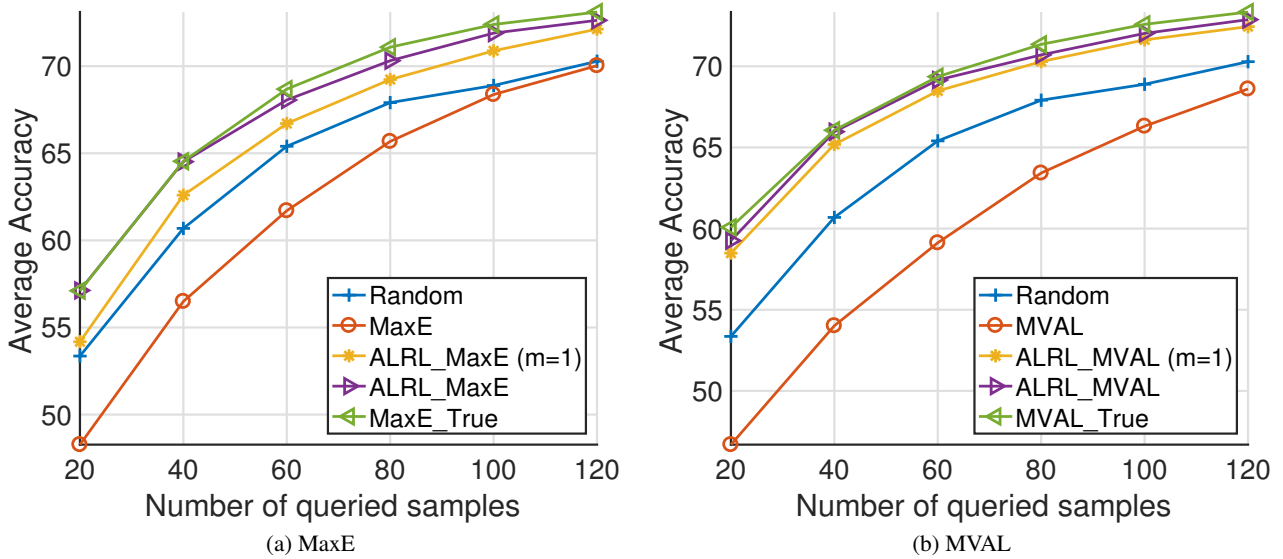

*Figure 4.* Performance comparisons of random labeling in combination with two active learners (a) MaxE and (b) MVAL. ALRL_MaxE is the combination of MaxE and our method ALRL with a default $m=10$ pseudo annotators while ALRL_MaxE ($m=1$) means that we only use a single random annotator so that MNND is not utilized. It is the same setting with ALRL_MVAL and ALRL_MVAL ($m=1$). MaxE_True and MVAL_True refer to the two active learners that can obtain the true label from human expert when an unlabeled sample is selected.

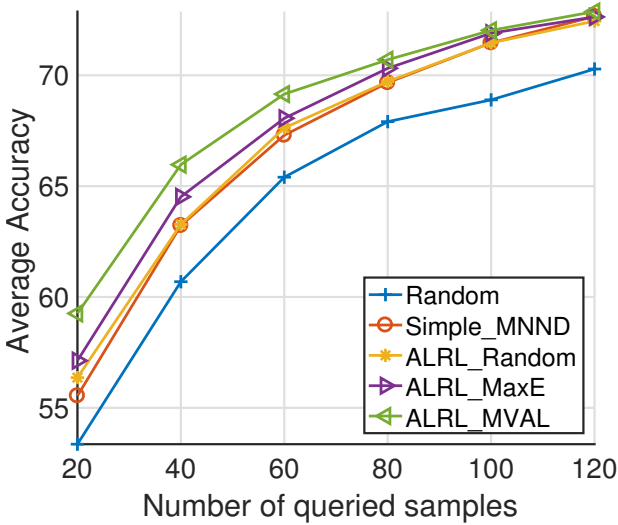

*Figure 5.* Performance comparisons of our method ALRL in combination with different active learners over all test sets. Simple_MNND means that we ignore the pseudo annotators and directly choose the most representative sample from all the remaining unlabeled samples by using MNND. ALRL_Random means that random sampling is used as the active learner $\mathcal{A}$ in Algorithm 1.

### 5.3.4. SENSITIVITY ANALYSIS

We also set up experiments to explore the influence of the number of pseudo annotators on the efficiency of our method. We use MaxE as the active learner and repeat the experiments by varying $m$ from a set $\{1, 4, 8, 10, 12, 16, 20, 24\}$. Figure 8 shows the average performance over 10 test datasets. Note that $m = 1$ means that no MNND is utilized since there is only a single candidate in $\mathcal{S}$. In the case of $m = 1$, a sharp decline in the performance of our proposed method is witnessed, which indicates that MNND can indeed enhance the performance by filtering out some poor random labelings. We can also observe that our method is not very sensitive to the number of pseudo annotators. For example, when $m$ varies from 4 to 24, the overall average performance of the proposed method shows little change. This implies that our method is robust to the number of pseudo annotators.

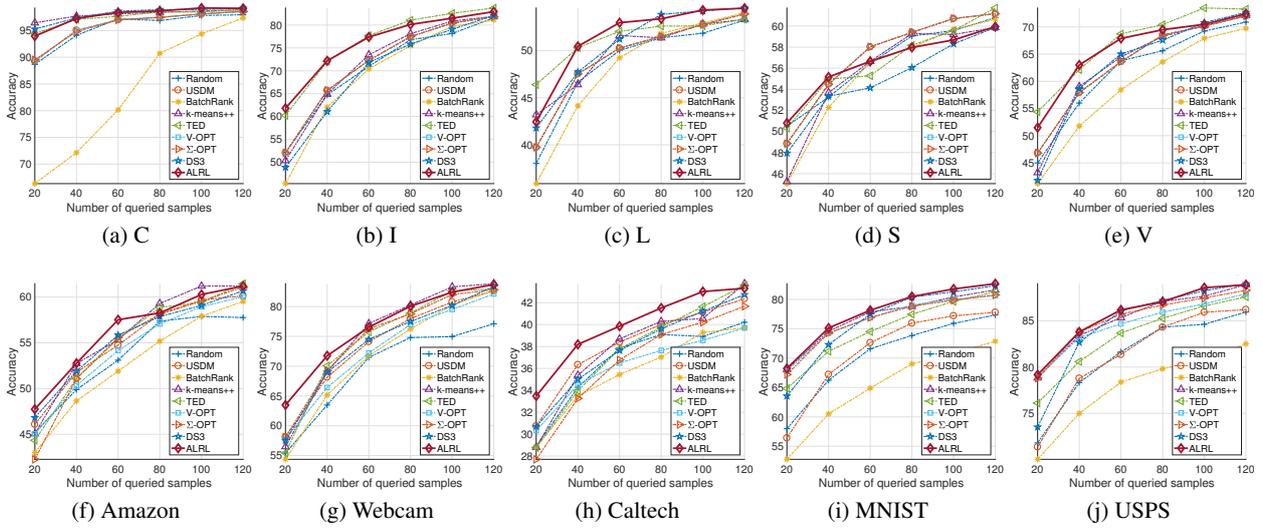

Figure 6. Performance comparison of different methods over 10 test datasets. The x-axis is the number of queried samples while the y-axis is the classification accuracy.

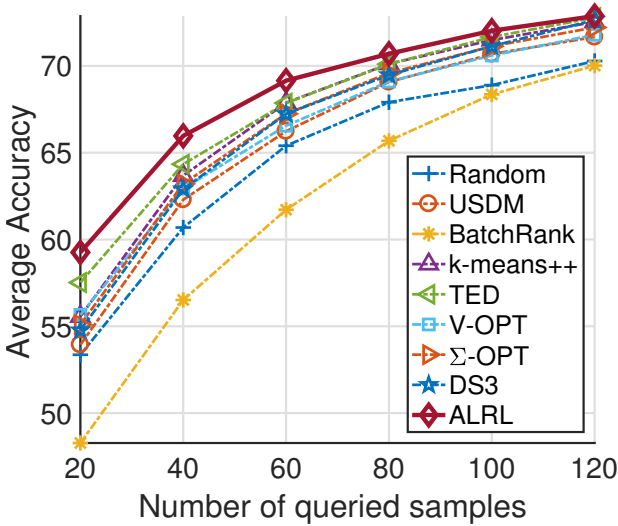

Figure 7. Overall average performance of compared methods on 10 test datasets.

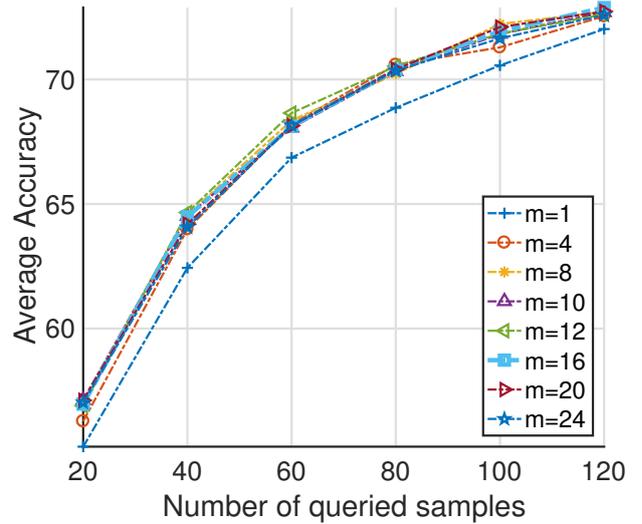

Figure 8. Sensitivity analysis: it shows average performance of our proposed method over 10 test datasets w.r.t. different number of pseudo annotators $m$.

## 6. Conclusion

We tackle the problem of human experts being unavailable during the active data selection process by introducing multiple pseudo annotators. These pseudo annotators uniformly and randomly annotate queried samples, which provides standard active learning methods with the ability to explore. The exploratory behavior is further enhanced by selecting the most representative sample through minimizing nearest neighbor distance between unlabeled data and queried data. Experiments on real-world datasets show that our method (ALRL) can indeed obtain close result to active learner that receives true label feedback from human annotator. Our method demonstrates very good performance when compared with state-of-the-art data selection methods.

## References

Arthur, David and Vassilvitskii, Sergei. k-means++: The advantages of careful seeding. In *Proceedings of the*

*eighteenth annual ACM-SIAM symposium on Discrete algorithms*, pp. 1027–1035. Society for Industrial and Applied Mathematics, 2007.

Boutsidis, Christos, Mahoney, Michael W, and Drineas, Petros. An improved approximation algorithm for the column subset selection problem. In *Proceedings of the twentieth annual ACM-SIAM symposium on Discrete algorithms*, pp. 968–977. Society for Industrial and Applied Mathematics, 2009.

Brinker, Klaus. Incorporating diversity in active learning with support vector machines. In *Proceedings of the 20th International Conference on Machine Learning (ICML-03)*, pp. 59–66, 2003.

Chakraborty, Shayok, Balasubramanian, Vineeth, and Panchanathan, Sethuraman. Generalized batch mode active learning for face-based biometric recognition. *Pattern Recognition*, 46(2):497–508, 2013.

Chakraborty, Shayok, Balasubramanian, Vineeth, Sun, Qian, Panchanathan, Sethuraman, and Ye, Jieping. Active batch selection via convex relaxations with guaranteed solution bounds. *IEEE Transactions on Pattern Analysis and Machine Intelligence*, 37(10):1945–1958, 2015.

Chang, Chih-Chung and Lin, Chih-Jen. Libsvm: a library for support vector machines. *ACM Transactions on Intelligent Systems and Technology (TIST)*, 2(3):27:1–27:27, 2011.

Contardo, Gabriella, Denoyer, Ludovic, and Artieres, Thierry. A meta-learning approach to one-step active learning. *arXiv preprint arXiv:1706.08334*, 2017.

Elhamifar, Ehsan, Sapiro, Guillermo, and Vidal, Rene. See all by looking at a few: Sparse modeling for finding representative objects. In *Computer Vision and Pattern Recognition (CVPR), 2012 IEEE Conference on*, pp. 1600–1607. IEEE, 2012.

Elhamifar, Ehsan, Sapiro, Guillermo, and Sastry, S Shankar. Dissimilarity-based sparse subset selection. *IEEE transactions on pattern analysis and machine intelligence*, 38(11):2182–2197, 2016.

Esteva, Andre, Kuprel, Brett, Novoa, Roberto A, Ko, Justin, Swetter, Susan M, Blau, Helen M, and Thrun, Sebastian. Dermatologist-level classification of skin cancer with deep neural networks. *Nature*, 542(7639):115–118, 2017.

Fang, Chen, Xu, Ye, and Rockmore, Daniel N. Unbiased metric learning: On the utilization of multiple datasets and web images for softening bias. In *International Conference on Computer Vision*, 2013.

Frey, Brendan J and Dueck, Delbert. Clustering by passing messages between data points. *science*, 315(5814):972–976, 2007.

Freytag, Alexander, Rodner, Erik, and Denzler, Joachim. Selecting influential examples: Active learning with expected model output changes. In *Computer Vision–ECCV 2014*, pp. 562–577. Springer, 2014.

Gong, Boqing, Shi, Yuan, Sha, Fei, and Grauman, Kristen. Geodesic flow kernel for unsupervised domain adaptation. In *Computer Vision and Pattern Recognition (CVPR), 2012 IEEE Conference on*, pp. 2066–2073. IEEE, 2012.

Guo, Yuhong and Schuurmans, Dale. Discriminative batch mode active learning. In *Proceedings of the 20th International Conference on Neural Information Processing Systems*, pp. 593–600, 2007.

Hoi, Steven CH, Jin, Rong, Zhu, Jianke, and Lyu, Michael R. Batch mode active learning and its application to medical image classification. In *Proceedings of the 23rd International Conference on Machine Learning*, pp. 417–424. ACM, 2006.

Hull, Jonathan J. A database for handwritten text recognition research. *IEEE Transactions on Pattern Analysis and Machine Intelligence*, 16(5):550–554, 1994.

Ji, Ming and Han, Jiawei. A variance minimization criterion to active learning on graphs. In *Artificial Intelligence and Statistics*, pp. 556–564, 2012.

LeCun, Yann, Bottou, Léon, Bengio, Yoshua, and Haffner, Patrick. Gradient-based learning applied to document recognition. *Proceedings of the IEEE*, 86(11):2278–2324, 1998.

Lewis, David D and Gale, William A. A sequential algorithm for training text classifiers. In *Proceedings of the 17th annual international ACM SIGIR conference on Research and development in information retrieval*, pp. 3–12, 1994.

Li, Chun-Liang, Ferng, Chun-Sung, and Lin, Hsuan-Tien. Active learning using hint information. *Neural computation*, 27(8):1738–1765, 2015.

Liu, Rujie, Wang, Yuehong, Baba, Takayuki, Masumoto, Daiki, and Nagata, Shigemi. Svm-based active feedback in image retrieval using clustering and unlabeled data. *Pattern Recognition*, 41(8):2645–2655, 2008.

Lu, Ke, Zhao, Jidong, and Wu, Yue. Hessian optimal design for image retrieval. *Pattern Recognition*, 44(6):1155–1161, 2011.

Ma, Yifei, Garnett, Roman, and Schneider, Jeff. $\sigma$-optimality for active learning on gaussian random fields. In *Advances in Neural Information Processing Systems*, pp. 2751–2759, 2013.

Roy, Nicholas and Mccallum, Andrew. Toward optimal active learning through sampling estimation of error reduction. In *In Proc. 18th International Conf. on Machine Learning*, pp. 441–448, 2001.

Schein, Andrew I and Ungar, Lyle H. Active learning for logistic regression: an evaluation. *Machine Learning*, 68(3):235–265, 2007.

Settles, Burr. Active learning literature survey. *University of Wisconsin, Madison*, 52(55-66):11, 2010.

Settles, Burr and Craven, Mark. An analysis of active learning strategies for sequence labeling tasks. In *Proceedings of the Conference on Empirical Methods in Natural Language Processing*, pp. 1070–1079. Association for Computational Linguistics, 2008.

Seung, H. S., Opper, M., and Sompolinsky, H. Query by committee. In *Proceedings of the Fifth Annual Workshop on Computational Learning Theory*, COLT '92, pp. 287–294, 1992.

Tong, Simon and Koller, Daphne. Support vector machine active learning with applications to text classification. *The Journal of Machine Learning Research*, 2:45–66, 2002.

Wang, Meng and Hua, Xian-Sheng. Active learning in multimedia annotation and retrieval: A survey. *ACM Transactions on Intelligent Systems and Technology (TIST)*, 2(2):10, 2011.

Yang, Yazhou and Loog, Marco. A benchmark and comparison of active learning methods for logistic regression. *arXiv preprint*, 2016.

Yang, Yazhou and Loog, Marco. A variance maximization criterion for active learning. *Pattern Recognition*, 78:358–370, 2018.

Yang, Yi, Ma, Zhigang, Nie, Feiping, Chang, Xiaojun, and Hauptmann, Alexander G. Multi-class active learning by uncertainty sampling with diversity maximization. *International Journal of Computer Vision*, 113(2):113–127, 2015.

Yu, Kai, Bi, Jinbo, and Tresp, Volker. Active learning via transductive experimental design. In *Proceedings of the 23rd International Conference on Machine Learning*, pp. 1081–1088. ACM, 2006.

Zhu, Jingbo, Wang, Huizhen, Tsou, Benjamin K, and Ma, Matthew. Active learning with sampling by uncertainty and density for data annotations. *Audio, Speech, and Language Processing, IEEE Transactions on*, 18(6):1323–1331, 2010.